\pdfoutput=1

\documentclass[11pt]{article}

\usepackage[]{EMNLP2022}

\usepackage{times}
\usepackage{latexsym}
\usepackage{graphicx} 
\usepackage{subfigure}
\usepackage{amssymb,amsthm,dsfont,mathrsfs}
\usepackage{amsmath}
\usepackage{multirow}
\usepackage{mathrsfs}
\usepackage{float}
\usepackage{hyperref}

\usepackage[linesnumbered, ruled]{algorithm2e}

\def\<{\left\langle }
\def\>{\right\rangle }

\def\digamma{{\mathrm{digamma}}}

\usepackage[T1]{fontenc}

\usepackage[utf8]{inputenc}

\usepackage{microtype}

\usepackage{inconsolata}
\title{Efficient Adversarial Training with Robust Early-Bird Tickets}

\author{{\normalsize
    Zhiheng Xi$^1$\thanks{~~Equal contribution.}\ \ , \ \ Rui Zheng$^{1*}$, \ \  \textbf{Tao Gui}$^{2}$\thanks{{ }{ }Corresponding author.}\ \ \textbf{,} \ \ \textbf{Qi Zhang}$^{1}$\textbf{,} \ \ \textbf{Xuanjing Huang}$^{1,3}$ } 
  \\
  {$^1$ \normalsize School of Computer Science, Fudan University, Shanghai, China} \\
  {$^2$ \normalsize Institute of Modern Languages and Linguistics, Fudan University, Shanghai, China} \\
  {$^3$ \normalsize International Human Phenome Institutes (Shanghai), Shanghai, China} \\
  \texttt{\normalsize zhxi22@m.fudan.edu.cn} \texttt{,} \texttt{\normalsize \{rzheng20,tgui,qz,xjhuang\}@fudan.edu.cn}\\
  }

\begin{document}
\maketitle
\begin{abstract}
Adversarial training is one of the most powerful methods to improve the robustness of pre-trained language models (PLMs).
However, this approach is typically more expensive than traditional fine-tuning because of the necessity to generate adversarial examples via gradient descent. 
Delving into the optimization process of adversarial training, we find that robust connectivity patterns emerge in the early training phase (typically $0.15\sim0.3$ epochs), far before parameters converge.
Inspired by this finding, we dig out robust early-bird tickets (i.e., subnetworks) to develop an efficient adversarial training method: (1) searching for robust tickets with structured sparsity in the early stage;
(2) fine-tuning robust tickets in the remaining time.
To extract the robust tickets as early as possible, we design a ticket convergence metric to automatically terminate the searching process.
Experiments show that the proposed efficient adversarial training method can achieve up to $7\times \sim 13 \times$ training speedups while maintaining comparable or even better robustness compared to the most competitive state-of-the-art adversarial training methods.

\end{abstract}

\section{Introduction}

Pre-trained language models (PLMs) have achieved great success in NLP \cite{devlin-etal-2019-bert}, but they are vulnerable to adversarial examples crafted by performing subtle perturbations on normal examples \cite{DBLP:conf/acl/RenDHC19,garg-ramakrishnan-2020-bae}.
\begin{figure}[h]
    \includegraphics[width=0.96\linewidth]{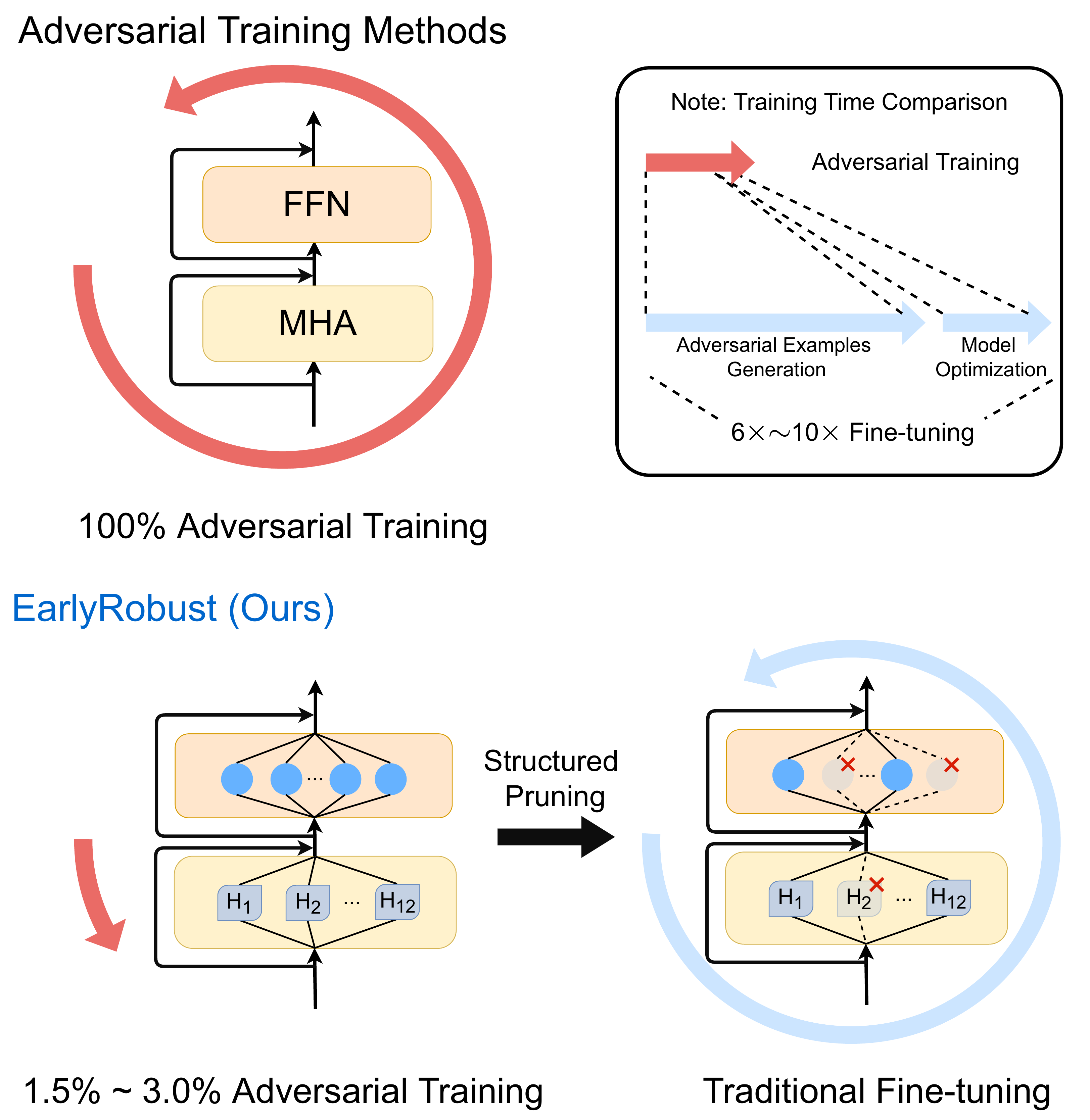}
    \centering
	\caption{A high-level overview of adversarial training methods and the proposed EarlyRobust. \textbf{MHA} means multi-head attention and \textbf{FFN} means feed-forward network. \textbf{Structured pruning} removes attention heads and intermediate neurons of each transformer layer in BERT. Our method achieves up to $7\times \sim 13\times$ training speedups compared to adversarial training methods.
	}
	\label{fig:Overview}
\end{figure}
Recent studies have shown the prevalence of adversarial vulnerability in NLP tasks \cite{wallace-etal-2019-trick,zhang-etal-2021-crafting,lin-etal-2021-using}.
Various defense strategies have been proposed to improve the robustness of the model and maintain high accuracy on both normal and adversarial examples \cite{DBLP:conf/iclr/ZhuCGSGL20,DBLP:conf/iclr/WangWCGJLL21}.
Adversarial training is one of the most widely used and strongest defense methods \cite{DBLP:conf/iclr/MadryMSTV18}.

One of the main challenges of adversarial training in real-world applications is its high computational cost, as they require multi-step gradient descents to generate adversarial examples \cite{DBLP:conf/iclr/WongRK20,DBLP:conf/nips/AndriushchenkoF20}.
Some work on this topic found that replacing strong adversarial examples with weaker and cheaper ones does not significantly change the resulting robustness \cite{DBLP:conf/iclr/WongRK20}.
\citet{DBLP:conf/nips/ZhangZLZ019} remove redundant calculations during backpropagation for additional speedup.
 FreeAT \cite{DBLP:conf/nips/ShafahiNG0DSDTG19} and FreeLB \cite{DBLP:conf/iclr/ZhuCGSGL20} utilize  a ``free'' strategy to generate diverse adversarial examples at a negligible additional cost.
However, these methods are only alleviating the huge expense of generating adversarial samples, 
and still require optimizing the adversarial loss objective throughout the training period.

The recently proposed robust lottery ticket hypothesis \cite{DBLP:conf/nips/FuYZWOCL21,zheng-etal-2022-robust} suggests that a deep network contains robust tickets (i.e., subnetworks) that maintain matching accuracy but better robustness than the original network when trained individually.
However, robust tickets have to be searched through a tedious process under the guidance of the adversarial loss objective \cite{zheng-etal-2022-robust}.
Moreover, these robust tickets consider unstructured sparsity, which only reduces storage but does not reduce computational overhead \cite{DBLP:conf/emnlp/PrasannaRR20}.

In this work, we propose a novel \textbf{robust} \textbf{early-bird} ticket with \textbf{structured sparsity} for efficient adversarial training. We delve into the optimization process of adversarial training, and find that the robust connectivity patterns converge with few training iterations.
As shown in Figure \ref{fig:Overview}, we can extract robust tickets at the early stage of adversarial training by pruning the self-attention heads and intermediate neurons that contribute least to accuracy and robustness. 
A highly robust model can be easily obtained by fine-tuning the proposed robust tickets in the remaining time.
In addition, we use a generic ticket convergence metric to automatically terminate the search process without going through each search moment.
Experimental results show that the proposed method achieves up to $7\times \sim 13 \times$ training speedups while maintaining comparable or even better adversarial robustness compared with traditional adversarial training. Our codes are publicly available at \textit{Github}\footnote{\href{https://github.com/WooooDyy/EarlyRobust}{https://github.com/WooooDyy/EarlyRobust}}.
The contributions of this work can be summarized as:
\begin{itemize}
\setlength{\itemindent}{0em}
\setlength{\itemsep}{0em}
\setlength{\topsep}{-0.5em}

\item We analyze the optimization process of adversarial training and reveal that the robust connectivity patterns emerge in the early training phase. 
    
\item We propose a novel robust early-bird ticket for efficient adversarial training, which consists of two stages: (1) searching for robust tickets using adversarial training in the early stage; and (2) fine-tuning the robust tickets during the remaining time.

\item Compared with previous efficient adversarial training methods, the proposed approach provides a new pathway based on model pruning and robust architecture searching.

\end{itemize}

\section{Related Work}
\subsection{Textual Adversarial Attack and Defense}
Textual adversarial attacks try to fool NLP models with adversarial examples which are constructed by substituting some parts of sentences with their counterparts. Typically, the adversarial examples hold a high similarity with clean ones in semantics or embedding space \cite{DBLP:conf/ndss/LiJDLW19,DBLP:conf/acl/RenDHC19,DBLP:conf/aaai/JinJZS20,DBLP:conf/emnlp/LiMGXQ20}. To improve the robustness of NLP models against textual adversarial attacks, many defense methods have been proposed \cite{DBLP:conf/emnlp/LiXZLZZCH21,DBLP:conf/coling/ZhengBLGZHXW22,DBLP:conf/acl/LiuZRLLCQG0H22}. As the most popular one, adversarial training (AT) solves a robust min-max optimization problem by adding norm-bounded perturbations to word embeddings \cite{DBLP:conf/iclr/MadryMSTV18,DBLP:conf/iclr/ZhuCGSGL20,DBLP:conf/aaai/LiQ21}. Likewise, some regularization methods have been proved beneficial to model robustness \cite{DBLP:conf/acl/JiangHCLGZ20,DBLP:conf/iclr/WangWCGJLL21}.

\subsection{Efficient Adversarial Training Methods}
Adversarial training, represented by PGD (Projected Gradient Descent, \citet{DBLP:conf/iclr/MadryMSTV18}), costs much more computational resources than traditional training due to the need to construct adversarial examples. 
Therefore, some recent works attempt to improve its efficiency. FreeAT \cite{DBLP:conf/nips/ShafahiNG0DSDTG19} and YOPO \cite{DBLP:conf/nips/ZhangZLZ019} simplify the calculation of gradients to get speedups. \citet{DBLP:conf/iclr/ZhuCGSGL20} introduce the above accelerating methods to NLP and propose FreeLB, which adds adversarial perturbations to word embeddings. 
Our method shapes structured robust tickets in the early adversarial training stage, so it is orthogonal to the above methods.

\subsection{Lottery Ticket Hypothesis}
Lottery Ticket Hypothesis (LTH) suggests that there exist certain sparse subnetworks (i.e., winning tickets) at initialization that can be trained to achieve competitive performance compared to the full model \cite{DBLP:conf/iclr/FrankleC19}. In NLP, previous work has found that matching subnetworks exist in Transformers, LSTMs, and PLMs \cite{DBLP:conf/iclr/YuETM20,DBLP:conf/iclr/RendaFC20,DBLP:conf/nips/ChenFC0ZWC20}. 
\citet{DBLP:conf/emnlp/PrasannaRR20} find structured tickets for BERT, which are efficient in training and inference, by pruning attention heads and MLP layers. 
Recently, \citet{DBLP:conf/iclr/YouL0FWCBWL20} and \citet{DBLP:conf/acl/ChenCWGWL20} pioneer to identify Early-Bird tickets at the early stage of training.
\citet{DBLP:conf/nips/FuYZWOCL21} and \citet{zheng-etal-2022-robust} find robust tickets in neural networks and thus the robust LTH is proposed. However, the search process for winning robust tickets is tedious and they perform unstructured pruning. 
Inspired by them, we extract robust early-bird tickets with structured sparsity, which can be fine-tuned to have comparable or even better robustness. 
A concurrent work \cite{DBLP:conf/iclr/ChenZWBMWW22} discusses similar findings in CV under unstructured sparsity, while we explore it in NLP under structured sparsity.
\section{Methodology}

\begin{figure*}[htbp]
	\centering
	\subfigure[IMDB MHA]{
        \begin{minipage}[t]{0.3\linewidth}
        \centering
        \includegraphics[width=0.9\linewidth]{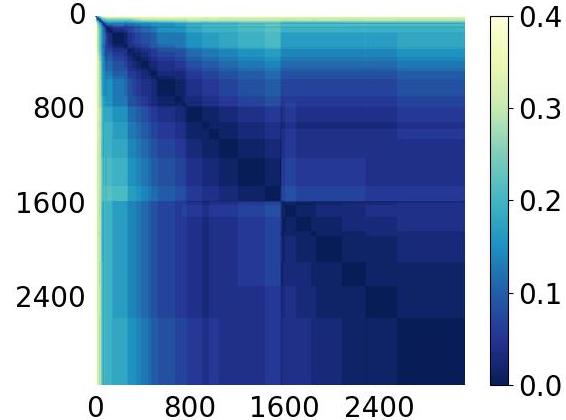}
        \label{mask_dis_imdb_self}
        \end{minipage}%
    }%
	\centering
	\subfigure[SST-2 MHA]{
        \begin{minipage}[t]{0.3\linewidth}
        \centering
        \includegraphics[width=0.9\linewidth]{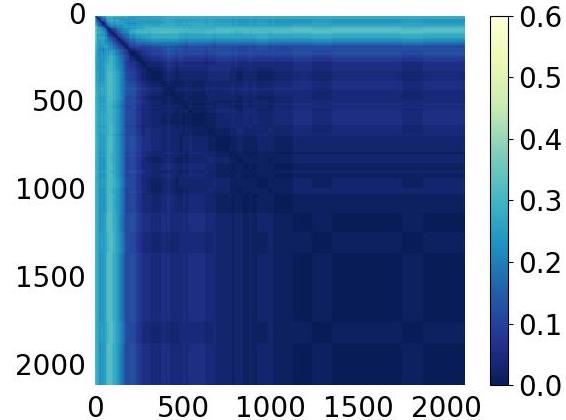}
        \label{mask_dis_sst2_self}
        \end{minipage}%
    }%
	\centering
	\subfigure[AG NEWS MHA]{
        \begin{minipage}[t]{0.3\linewidth}
        \centering
        \includegraphics[width=0.9\linewidth]{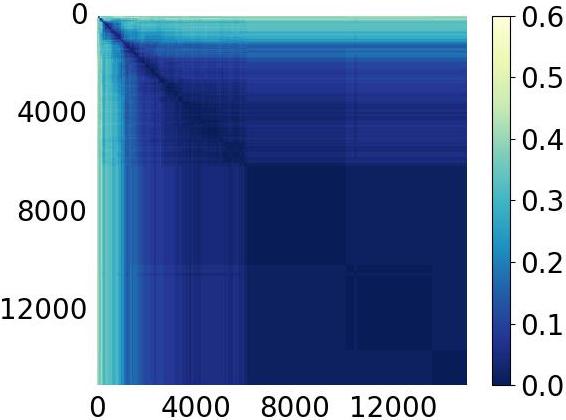}
        \label{mask_dis_ag_news_self}
        \end{minipage}%
    }%
	
	\centering
	\subfigure[IMDB FFN]{
        \begin{minipage}[t]{0.3\linewidth}
        \centering
        \includegraphics[width=0.9\linewidth]{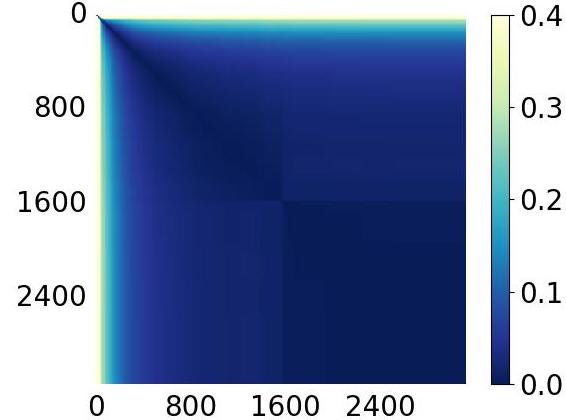}
        \label{imdb inter}
        \end{minipage}%
    }%
	\centering
	\subfigure[SST-2 FFN]{
        \begin{minipage}[t]{0.3\linewidth}
        \centering
        \includegraphics[width=0.9\linewidth]{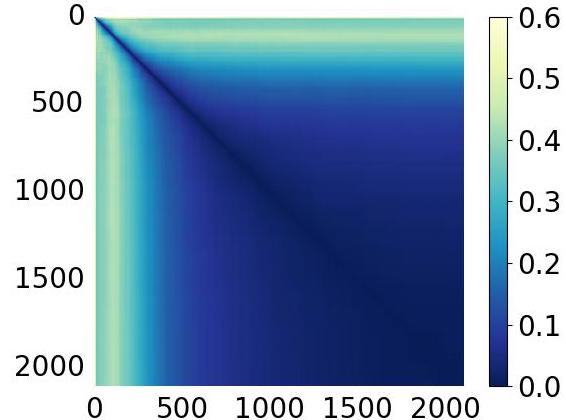}
        \label{mask_dis_sst2_inter}
        \end{minipage}%
    }%
	\centering
	\subfigure[AG NEWS FFN]{
        \begin{minipage}[t]{0.3\linewidth}
        \centering
        \includegraphics[width=0.9\linewidth]{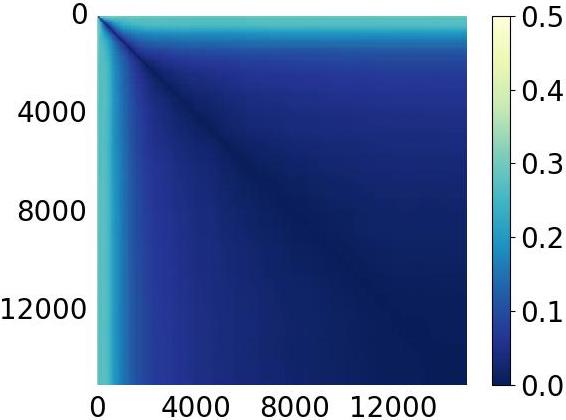}
        \label{mask_dis_ag_news_inter}
        \end{minipage}%
    }%

	\caption{Visualization of mask difference in Hamming distance between different training steps. Top: mask distance in self-attention heads. Bottom: mask distance in intermediate neurons. The axes represent the numbers of \textbf{training steps} finished and the color stands for the \textbf{normalized distance} between different masks. The darker the color, the smaller the distance. The masks for attention heads and intermediate neurons both converge early.    }
	\label{fig:Normed Mask Distance}
\end{figure*}

In this section, we first revisit the robust lottery ticket hypothesis (Sec.\ref{sec:Revisiting Robust Lottery Ticket Hypothesis}). Next, we describe the early emergence of structured robust tickets (Sec.\ref{Sec:The Early Emergency of Structured Robust Tickets}). Based on this key finding, we propose the EarlyRobust method for efficient adversarial training (Sec.\ref{Sec:Discovering Structured Robust Early-bird Tickets}). Finally, we make a brief review and discuss how our method accelerates adversarial training (Sec.\ref{sec:A Brief Review}).

\subsection{Revisiting Robust Lottery Ticket Hypothesis} \label{sec:Revisiting Robust Lottery Ticket Hypothesis}

For a network $f(\theta_0)$ initialized with parameters $\theta_0$, we can identify a binary mask $m$ which has the same dimension as $\theta_0$, such that a subnetwork $f(m\odot \theta_0)$\footnote{$\odot$ is the Hadamard product operator.} can be shaped. Robust lottery ticket hypothesis articulates that, there exists certain subnetwork that can be trained to have matching performance and better robustness compared to the full model following the same training protocol \cite{zheng-etal-2022-robust}. 
Such a subnetwork $f(m\odot \theta_0)$ is a so-called \emph{robust winning ticket}, including both the mask $m$ and the initialization $\theta_0$. 

\subsection{Early Emergence of Robust Tickets with Structured Sparsity}\label{Sec:The Early Emergency of Structured Robust Tickets}

In previous work, robust tickets with unstructured sparsity in PLMs are extracted after a tedious training process.
However, when applying a structured pruning strategy, we observe that \textbf{robust tickets with structured sparsity emerge in the early training stage.} 
The structured pruning strategy aims at pruning self-attention heads and intermediate neurons of each transformer layer in BERT.
Through visualizing the normalized mask distance between different training steps \cite{DBLP:conf/iclr/YouL0FWCBWL20} in Figure \ref{fig:Normed Mask Distance}, we learn that the structure of a robust ticket can be identified at a very early adversarial training stage.
Inspired by this finding, we propose an efficient adversarial training method that draws robust early-bird tickets at the early adversarial training stage and then fine-tunes these tickets to achieve better robustness.

\subsection{Robust Early-bird Tickets}\label{Sec:Discovering Structured Robust Early-bird Tickets}
Our method consists of two steps: 1) Searching Stage; 2) Drawing and Fine-tuning Stage.

\subsubsection{Searching Stage} \label{sec:Searching for Robust Talent}
BERT is constructed by a stack of transformer encoder layers \cite{DBLP:conf/nips/VaswaniSPUJGKP17}, and the structure of each layer is identical: a multi-head self-attention (MHA) block followed by a feed-forward network (FFN), with residual connections around each. 
In each layer, the MHA with $N_h$ heads takes an input $x$ and outputs:
\begin{equation}
\begin{aligned}
{\rm MHA}(x) = \sum_{i=1}^{N_h}{{\rm Att}_{W_K^{i},W_Q^{i},W_V^{i},W_O^{i} }(x)},
\label{eqn: Multi-head attention}
\end{aligned}
\end{equation}
where $W_K^{i}$, $W_Q^{i}$, $W_V^{i}$ $\in$ $\mathbb{R}^{d_h \times d}$, $W_O^{i} \in \mathbb{R}^{d\times d_h}$, and they denote the query, key, value and output matrices in the $l$-th attention head. Here $d$ denotes the hidden size (e.g., 768) and $d_h = d/N_h$ denotes the output dimension of each head (e.g., 64).

Next comes an FFN parameterized by $W_U\in \mathbb{R}^{d \times d_f}$ and $W_D\in \mathbb{R}^{d_f \times d}$:
\begin{equation}
    \begin{aligned}
    {\rm FFN}(x) = {\rm gelu}(XW_U)\cdot W_D,
    \end{aligned}
\end{equation}
where $d_f=4d$. 

\noindent \textbf{Learnable Importance Coefficients} Recent research reveals that MHAs and FFNs in transformers are redundant \cite{DBLP:conf/nips/MichelLN19,DBLP:conf/acl/VoitaTMST19}. 
Therefore, we adopt the modified network slimming \cite{DBLP:conf/iccv/LiuLSHYZ17,DBLP:conf/acl/ChenCWGWL20}, which assigns learnable importance coefficients to self-attention heads and intermediate neurons, to determine which components are essential for robustness:
\begin{align}
& {\rm MHA}(x) = \sum_{i=1}^{N_h}{c_{\rm  \scriptscriptstyle  H}^{i} \cdot {\rm Att}_{W_K^{i},W_Q^{i},W_V^{i},W_O^{i}}(x)}, \label{eqn: Multi-head attention} \\
&{\rm FFN}(x) = c_{{\rm \scriptscriptstyle F}} \cdot {\rm gelu}(XW_U)\cdot W_D,
\label{eqn: FFN with coef}
\end{align}
where $c_{\rm  \scriptscriptstyle  H}$ denotes the coefficients for heads, $i$ is the index of head, and $c_{{\rm \scriptscriptstyle F}}$ denotes the coefficients for the FFN. Then, we can jointly train BERT with the importance coefficients but with a regularizer:
\begin{equation}
\begin{aligned}
\mathcal R(c) = \lambda_{\rm  \scriptscriptstyle H} \lVert c_{\rm  \scriptscriptstyle H} \rVert_1 + \lambda_{{\rm \scriptscriptstyle F}} \lVert c_{{\rm \scriptscriptstyle F}} \rVert_1 ,\label{coef_regularizer}
\end{aligned}
\end{equation}
where $c=\{c_{\rm  \scriptscriptstyle H},  c_{{\rm \scriptscriptstyle F}}\}$, $\lambda_{\rm  \scriptscriptstyle H}$ and $\lambda_{{\rm \scriptscriptstyle F}}$ denote regularization strength for the two kinds of coefficients respectively.

\noindent \textbf{Adversarial Loss Objective} 
To identify substructures that are responsible for adversarial robustness, we introduce the adversarial loss objective:
\begin{equation}
\begin{aligned}
\min_{\theta,c} \underbrace{\mathbb{E}_{(x,y)\sim\mathcal{D}}  \max_{\lVert \delta \rVert  \leq \epsilon} \mathcal{L}\left(f(x+\delta;  \theta,c),y\right)}_{\mathcal{L}_{adv}(\theta,c)} \label{adv_loss_with_c},
\end{aligned}
\end{equation}
where $(x,y)$ is a data point from dataset $\mathcal{D}$, $\delta$ is the perturbation that is constrained within the $\epsilon$ ball. The inner max problem tries to generate worst-case perturbation towards inputs and maximize the model's classification loss, while the outer min problem is to optimize the model on the perturbed data.\footnote{See more details about how to solve the inner max problem in Appendix \ref{appendix:Solving the Inner Max Problem of Adversarial Loss Objective}.} 

Then, by integrating the adversarial training objective and the sparsity regularizer $\mathcal{R}(c)$, we can describe our loss objective as follow:
\begin{equation}
\begin{aligned}
    \min_{\theta,c} \mathcal{L}_{adv}(\theta,c) + \mathcal R(c) . \label{stage_one_obj}
\end{aligned}
\end{equation}

After the joint training, we draw robust tickets, reset the parameters and then train the tickets with traditional fine-tuning.

\subsubsection{Drawing and Fine-tuning Stage} \label{sec:Drawing and Fine-tuning Structured Robust tickets}
\noindent \textbf{Early-stopping Strategy}
As mentioned before, robust tickets with structured sparsity can be identified in the early adversarial training stage. However, it is difficult to determine the exact search termination time. The difficulty lies mainly in the following two aspects:
(1) the termination time varies among datasets; (2) the termination moments of MHA and FFN in the same model are different.
Therefore, we design a termination metric that measures the normalized mask distance between consecutive miniepochs (each miniepoch consists of $0.05$ epochs).
When the termination metric is smaller than the threshold $\gamma$ both for MHA and FFN, we extract robust early-bird tickets.
Similar termination metrics are widely used for the extraction of subnetworks \cite{DBLP:conf/iclr/YouL0FWCBWL20}.
To ensure the consistency of convergence, we end the subnetwork search stage when the termination metric is satisfied five times in a row.

\noindent \textbf{Pruning Strategy}
At the end of the search process, we prune the unimportant parts according to the magnitudes of the learned importance coefficients.
The attention heads and intermediate neurons with the smallest importance coefficients are considered to contribute the least to robustness and should be removed.
For attention heads, we not only remove them from the computation graph, but also remove the corresponding $W_O$ (see Eqn. (\ref{eqn: Multi-head attention})).
We adopt a modified global approach to prune the heads and ensure that at least one head survives in each layer.\footnote{The attention heads in some layers may be completely removed if we use a global pruning approach, leaving the network un-trainable. 
While our experiments (see Sec.\ref{sec:Global Pruning vs. Layer-wise Pruning}) show that global pruning is more effective than layer-wise pruning. The modified global approach is a trade-off between trainability and effectiveness. }
As for intermediate neurons, pruning them is equivalent to reducing the size of $W_U$ and $W_D$ in Eqn. (\ref{eqn: FFN with coef}). 
We prune intermediate neurons globally as there is a large number of neurons in each FFN layer and empirical analysis suggests that global approach brings better performance.

\noindent \textbf{Efficient Robust Fine-tuning} After pruning attention heads and intermediate neurons, we then reset the weights to the pre-trained weights and utilize traditional fine-tuning to train the robust tickets.

\subsection{A Brief Review} \label{sec:A Brief Review}
The proposed method searches for the key connectivity patterns for robustness
, draws the winning tickets and trains them with traditional fine-tuning. We summarize it in Algorithm \ref{alg:The EarlyRobust Method}.

\begin{algorithm}[t]
  \SetKwData{Left}{left}\SetKwData{This}{this}\SetKwData{Up}{up}
  \SetKwFunction{Union}{Union}\SetKwFunction{FindCompress}{FindCompress}
  \SetKwInOut{Input}{Input}\SetKwInOut{Output}{Output}
   \SetKwProg{myproc}{Procedure}{}{}
   \KwIn{model parameters $\theta$, learnable importance coefficients $c$, learning rate $\eta$ and the fine-tuning epoch $N$.}
    
   \textbf{Procedure} \textsc{Robust Tickets Searching}{{
	Initialize $\theta$, $c$\;}
	
	\Repeat{the convergence condition in Sec.\ref{sec:Drawing and Fine-tuning Structured Robust tickets} is satisfied}{
		$\theta = \theta - \eta \nabla_{\theta}(\mathcal{L}_{adv}(\theta,c) + \mathcal R(c))$\; 
		$c = c-\eta \nabla_{c}(\mathcal{L}_{adv}(\theta,c) + \mathcal R(c))$\;
	}
}
     \textbf{Procedure} \textsc{Drawing and fine-tuning}{
        Extract robust winning tickets parameterized by $\theta_{ticket}$\;
        \For{$epoch \gets 1 ... N$}{
             $\theta_{ticket} \gets \theta - \eta\nabla_{\theta_{ticket}}\mathcal{L}_{CE}$\;
        }
    }
  \caption{EarlyRobust Method}
  \label{alg:The EarlyRobust Method}
\end{algorithm}



    




\noindent \textbf{How does EarlyRobust accelerate adversarial training?} Firstly, the method draws robust tickets with \emph{structured sparsity}, which reduces the size of models and speeds up training and inference. Secondly, the \emph{early-stopping strategy} prevents the tedious searching process for robust tickets. Finally, the method trains the robust early-bird tickets with \emph{traditional fine-tuning} instead of adversarial training methods, getting rid of the time-consuming process to construct adversarial examples.

\section{Experimental Settings}

\begin{table*}[ht]
\renewcommand\arraystretch{1}
\setlength\tabcolsep{6pt}
\centering
\scalebox{0.82}{
\begin{tabular}{c|l|c|c|c|cc|cc|cc}
\hline
\hline
\multicolumn{1}{c|}{\multirow{2}{*}{\textbf{Dataset}}} &
\multicolumn{1}{c|}{\multirow{2}{*}{\textbf{Method}}} &
\multicolumn{1}{c|}{\multirow{2}{*}{\textbf{Speedup}}} &
\multicolumn{1}{c|}{\multirow{2}{*}{\textbf{Params}}} &
\multicolumn{1}{c|}{\multirow{2}{*}{\textbf{Clean$\%$}}} &
\multicolumn{2}{c|}{\textbf{TextFooler}} & 
\multicolumn{2}{c|}{\textbf{TextBugger}} & 
\multicolumn{2}{c}{\textbf{BERT-Attack}} \\ \cline{6-11}
\multicolumn{1}{c|}{} & \multicolumn{1}{c|}{} & \multicolumn{1}{c|}{}& \multicolumn{1}{c|}{}& \multicolumn{1}{c|}{}& 
\multicolumn{1}{c}{\textbf{Aua$\%$}} &
\multicolumn{1}{c|}{\textbf{\#Query} } & 
\multicolumn{1}{c}{\textbf{Aua$\%$}} & 
\multicolumn{1}{c|}{\textbf{\#Query} } & 
\multicolumn{1}{c}{\textbf{Aua$\%$} } & 
\multicolumn{1}{c}{\textbf{\#Query} } \\ \cline{1-11}
\hline
\multirow{7}{*}{\textbf{IMDB}}
& Fine-tune$^{\dag}$ &$9.4\times$  & $85M$ & $92.05$ & $11.8$  & $922.4$ & $23.2$ & $695.2$ & $7.8$  & $1187.0$ \\ 
& EarlyBERT$^{\dag}$ &$\mathbf{\underline{15.6\times}}$ & $\mathbf{\underline{58M}}$ & $91.9$ & $3.4$  & $874.9$ & $12.6$  & $649.7$ & $0.8$  & $1099.9$ \\
&PGD$^{\ddag}$ &$1.0\times$ & $85M$ & $\mathbf{93.2}$ & $30.2$  & $1562.8$ & $41.6$  & $905.8$ & $21.8$  & $2114.6$ \\
&FreeLB$^{\ddag}$ &$1.2\times$ & $85M$ & $\mathbf{93.2}$ & $35.0$ & $1736.9$ & $53.0$  & $\mathbf{1110.9}$ & $29.0$  & $2588.8$ \\
&InfoBERT$^{\ddag}$ &$0.3\times$ & $85M$ & $\mathbf{\underline{93.3}}$ & $49.6$  & $1932.3$ & $53.8$  & $1070.4$ & $47.2$  & $3088.8$  \\
&RobusT$^{\ddag}$ &$0.4\times$ & $\mathbf{68M}$ & $91.8$ & $\mathbf{58.6}$  & $\mathbf{1994.7}$ & $\mathbf{63.6}$  & $\mathbf{\underline{1153.3}}$  & $\mathbf{58.0}$ &$\mathbf{3120.2}$ \\
&EarlyRobust&$\mathbf{12.6\times}$ & $\mathbf{\underline{58M}}$ & $91.8$ & $\mathbf{\underline{63.8}}$  & $\mathbf{\underline{2012.8}}$ & $\mathbf{\underline{63.8}}$ & $1107.1$   & $\mathbf{\underline{63.0}}$  & $\mathbf{\underline{3125.6}}$ \\
\hline
\multirow{7}{*}{\textbf{SST-2}}
&Fine-tune$^{\dag}$&$10.1\times$ & $85M$ & $92.2$ & $17.9$  & $123.4$ & $40.6$  & $53.4$ & $13.3$  & $158.3$ \\ 
&EarlyBERT$^{\dag}$ &$\mathbf{\underline{18.6\times}}$ & $\mathbf{63M}$ & $92.3$ & $13.4$  & $111.5$ & $37.7$  & $52.9$ & $9.9$  & $138.9$ \\
&PGD$^{\ddag}$&$1.0\times$ & $85M$ & $\mathbf{\underline{93.2}}$ & $18.1$  & $118.5$ & $44.2$ & $56.0$ & $13.4$  & $151.3$ \\
&FreeLB$^{\ddag}$&$1.4\times$ & $85M$ & $91.7$ & $\mathbf{\underline{29.4}}$  & $\mathbf{\underline{152.6}}$ & $\mathbf{\underline{49.7}}$  & $\mathbf{58.6}$ & $\mathbf{\underline{23.8}}$ & $\mathbf{174.7}$ \\
&InfoBERT$^{\ddag}$&$0.6\times$ & $85M$ & $92.4$ & $18.3$ & $126.8$ & $42.4$  & $54.6$ & $15.0$  & $160.4$ \\
&RobusT$^{\ddag}$&$0.4\times$ & $\mathbf{\underline{60M}}$ & $90.9$ & $28.6$  & $\mathbf{149.8}$ & $43.1$  & $53.9$ & $20.8$  & $169.2$  \\
&\textbf{EarlyRobust} &$\mathbf{13.7\times}$ & $\mathbf{63M}$ & $\mathbf{92.8}$ & $\mathbf{28.8}$  & $142.8$ & $\mathbf{46.3}$  & $\mathbf{\underline{61.9}}$  & $\mathbf{22.7}$  & $\mathbf{\underline{187.1}}$ \\
\hline
\multirow{7}{*}{\textbf{AG NEWS}}
&Fine-tune$^{\dag}$&$6.8\times$ & $85M$ & $94.6$ & $28.6$  & $383.3$ & $45.2$  & $192.5$ & $17.6$  & $556.0$ \\ 
&EarlyBERT$^{\dag}$ &$\mathbf{\underline{7.2\times}}$ & $\mathbf{69M}$ & $94.5$ & $32.8$  & $392.3$ & $46.6$  & $195.4$ & $21.2$  & $571.1$ \\
&PGD$^{\ddag}$&$1.0\times$ & $85M$ & $\mathbf{\underline{95.0}}$ & $\mathbf{36.8}$  & $414.9$ & $\mathbf{\underline{56.4}}$  & $201.8$ & $21.6$ & $616.1$ \\
&FreeLB$^{\ddag}$&$1.3\times$ & $85M$ & $\mathbf{\underline{95.0}}$ & $34.8$  & $408.5$ & $\mathbf{54.2}$  & $\mathbf{210.3}$ & $20.4$  & $596.2$ \\
&InfoBERT$^{\ddag}$&$0.4\times$ & $85M$ & $94.5$ & $33.8$  & $395.6$ & $49.6$  & $194.1$ & $\mathbf{23.4}$  & $\mathbf{618.9}$ \\
&RobusT$^{\ddag}$&$0.4\times$ & $\mathbf{\underline{60M}}$ & $\mathbf{94.9}$ & $35.2$  & $\mathbf{415.6}$ & $49.0$  & $206.9$ & $21.8$  & $617.5$  \\
&\textbf{EarlyRobust} &$\mathbf{7.0\times}$ & $\mathbf{69M}$ & $94.4$ & $\mathbf{\underline{41.0}}$  & $\mathbf{\underline{416.2}}$ & $50.0$  & $\mathbf{\underline{216.4}}$  & $\mathbf{\underline{32.2}}$  & $\mathbf{\underline{620.1}}$ \\
\hline
\hline
\end{tabular}
}
\caption{Experimental results of adversarial robustness evaluation. The best performance is marked in \textbf{bold} and \underline{underline}; the second is marked in \textbf{bold}. \textbf{Speedup} means training speedup, which is reported against adversarial training method PGD. \textbf{Params} is the number of model parameters.\footnotemark  
Methods labeled by  $\dag$ are fine-tuning baselines without considering adversarial robustness, and methods labeled by $\ddag$ are adversarial defense baselines. 
Our method achieves high adversarial robustness while maintaining a low computational and storage consumption. 
}

\label{tab:main results}
\end{table*}

\subsection{Backbone Model and Datasets}
We take BERT$_{BASE}$ (12 transformer layers, 12 attention heads, 3,072 intermediate neurons per layer, hidden size 768 and 110M parameters in total) as the backbone model. We follow the BERT implementation in \cite{DBLP:conf/naacl/DevlinCLT19,DBLP:journals/corr/abs-1910-03771}.
Mainly, we evaluate the proposed method on three text classification datasets, including IMDB \cite{DBLP:conf/acl/MaasDPHNP11}, SST-2 \cite{DBLP:conf/emnlp/SocherPWCMNP13}, and AG NEWs \cite{DBLP:conf/nips/ZhangZL15}. We also include other datasets of more tasks in GLUE \cite{DBLP:conf/iclr/WangSMHLB19} as a supplement, such as QNLI and QQP. 

\subsection{Baselines}
We compare our method with fine-tuned BERT, EarlyBERT, and several strong robust baselines.
\textbf{Fine-tune} \cite{DBLP:conf/naacl/DevlinCLT19}: Training full BERT on downstream tasks. \textbf{EarlyBERT} \cite{DBLP:conf/acl/ChenCWGWL20}: A framework that draws early-bird tickets from BERT for efficient fine-tuning. \textbf{PGD} \cite{DBLP:conf/iclr/MadryMSTV18}: An adversarial training algorithm that minimizes the empirical loss on worst-case adversarial examples. \textbf{FreeLB} \cite{DBLP:conf/iclr/ZhuCGSGL20}: An enhanced PGD-based adversarial training method for natural language understanding. \textbf{InfoBERT} \cite{DBLP:conf/iclr/WangWCGJLL21}: A framework that improves model's robustness from an information-theoretic perspective. \textbf{RobusT} \cite{zheng-etal-2022-robust}: An approach that identifies robust tickets from the original PLMs through learning binary masks.

\subsection{Evaluation Settings}  

\noindent \textbf{Robust Evaluation} We apply three popular textual attack methods \cite{DBLP:conf/emnlp/LiXZLZZCH21} to evaluate the model's defensive capability. \textbf{TextBugger} \cite{DBLP:conf/ndss/LiJDLW19} generates misspelled words by using character-level and word-level perturbations. 
\textbf{TextFooler} \cite{DBLP:conf/aaai/JinJZS20} constructs adversarial examples by substituting the most important words in a sentence with synonyms.
\textbf{BERT-Attack} \cite{DBLP:conf/emnlp/LiMGXQ20} employs BERT to generate adversarial texts, and thus the semantic consistency and the language fluency are preserved.

We adopt three evaluating metrics as follows: \textbf{Clean accuracy (Clean\%)} is the model's accuracy on the clean test set. \textbf{Accuracy under attack (Aua\%)} denotes the model's prediction accuracy under certain adversarial attack methods. \textbf{Number of Queries (\#Query)} refers to the average attempts an attacker queries the model, and the larger the number of queries is, the more difficult the model is to be fooled.

\noindent \textbf{Training Time Measurement Protocol} We measure the training time of each method on GPU and exclude the time for data I/O. To get rid of randomness, we run each method five times and report the average time. Notably, the training time of LTH-based methods (e.g., EarlyRobust, EarlyBERT, and RobusT) includes both the searching stage and the fine-tuning stage.

\noindent \subsection{Implementation Details} 
\footnotetext{We exclude embedding matrices when calculating the number of parameters following previous work. }

We re-implement baseline methods based on their open-source codes and the results are competitive. We employ FreeLB to implement the adversarial loss objective in the searching stage of EarlyRobust because compared to standard $K$-step PGD, it accumulates gradients of parameters in multiple forward passes and passes gradients backward once. \textbf{Clean\%} is tested on the whole test set. \textbf{Aua\%} and \textbf{\#Query} are evaluated on the whole test dataset for SST-2, and 500 randomly chosen samples for other datasets. We set the early-stopping threshold $\gamma$ to $0.1$ and fine-tune EarlyRobust tickets 10 epochs. For fine-tuning, EarlyBERT, and other adversarial training methods, we also set the training epoch to 10, which is a trade-off between training cost and performance. The sparsity for EarlyBERT is the same as that of EarlyRobust. We prune $40\%$, $30\%$, $20\%$, $30\%$ and $30\%$ intermediate neurons on IMDB, SST-2, AG NEWS, QNLI and QQP, respectively; the pruning ratio for attention heads is $1/6$. More implementation details and hyperparameters can be found in Appendix \ref{Appendix:Implementation Details}.
\section{Experimental Results and Discussion}
\begin{table}[t]
\renewcommand\arraystretch{1.1}
\setlength\tabcolsep{3pt}
\centering
\scalebox{0.8}{
\begin{tabular}{c|l|c|c|c|c}
\hline
\hline
\textbf{Dataset} &
\multicolumn{1}{c|}{\multirow{1}{*}{\textbf{Method}}} &
\textbf{Speedup} &
\textbf{Params} &
\textbf{Clean$\%$} &
\textbf{Aua$\%$} \\
\hline
\multirow{4}{*}{\textbf{QNLI}}
&Fine-tune$^{\dag}$ &$9.6\times$  & $85M$ & $\mathbf{\underline{91.6}}$ & $5.8$ \\
&PGD$^{\ddag}$ &$1.0\times$ & $85M$ & $91.2$ & $12.2$  \\
&FreeLB$^{\ddag}$ &$1.3\times$ & $85M$ & $90.5$ & $12.8$   \\
&\textbf{EarlyRobust} &$\mathbf{\underline{12.2\times}}$ & $\mathbf{\underline{63M}}$ & $91.4$ & $\mathbf{\underline{18.8}}$    \\
\hline
\multirow{4}{*}{\textbf{QQP}}
&Fine-tune$^{\dag}$  &$9.8\times$ & $85M$ & $\mathbf{\underline{91.3}}$ & $24.8$    \\ 
&PGD$^{\ddag}$ &$1.0\times$ & $85M$ & $91.2$ & $27.0$    \\
&FreeLB$^{\ddag}$  &$1.3\times$ & $85M$ & $91.2$ & $27.4$    \\
&\textbf{EarlyRobust} &$\mathbf{\underline{13.6\times}}$ & $\mathbf{\underline{63M}}$ & $90.9$ & $\mathbf{\underline{32.6}}$   \\
\hline
\hline
\end{tabular}
}
\caption{ Experimental results on QNLI and QQP. \textbf{Aua$\%$} is obtained after using TextFooler attack. Our method brings robustness improvement and training speedups on different tasks.
}
\label{tab:NLI datasets}
\end{table}

In this section, we illustrate the effectiveness and efficiency of our method with experimental results.
\subsection{Main Results}

The main results of EarlyRobust and other baselines are summarized in Table \ref{tab:main results}. We can observe that: 
1) Our method achieves high robustness with a little sacrifice in accuracy. It performs slightly worse than FreeLB on SST-2, but achieves the best robustness on both IMDB and AG NEWS, suggesting that the subnetworks shaped by our method own inborn robustness which can be exerted through traditional fine-tuning. 
2) Our method achieves sizable training accelerations compared to robust baselines. This proves our method can deliver highly robust models under different kinds of adversarial attacks with limited computational resources. Though EarlyBERT and fine-tuning are also fast, they do not take robustness into account and thus have weak defensive capabilities.
3) Our method also brings storage savings, and thus robust early-bird tickets are suitable to be deployed on mobile devices and edge devices.

We evaluate our method on paraphrase and inference tasks to verify its generality. Results in Table \ref{tab:NLI datasets} show that our method consistently performs well on more difficult tasks.

\subsection{Ablation Study}

To investigate the contribution of different components to our method, we perform ablation experiments by removing adversarial loss objective (Adv) and sparsity-inducing regularizer.
We also evaluate the performance of randomly pruned tickets. Results in Table \ref{tab:ablation} show that: 1) Adversarial loss objective is important for robustness, but not for clean accuracy.
2) There exists degradation in both accuracy and robustness if we remove the regularizer, indicating that the sparsity it induces is indispensable for high-quality tickets. 
3) Randomly pruned tickets have a competitive accuracy performance, which is also observed by \cite{DBLP:conf/emnlp/PrasannaRR20}. However, random tickets consistently suffer a large drop in robustness, which proves that robust tickets are non-trivial.

\begin{table}[t]
\renewcommand\arraystretch{1.15}
\setlength\tabcolsep{5pt}
\centering
\small
\begin{tabular}{l|lcc}
\hline
\hline
\textbf{Dataset}   & \multicolumn{1}{c}{\textbf{Method}} & \textbf{Clean$\%$} & \textbf{Aua$\%$} \\ \hline
 \multirow{4}{*}{\textbf{IMDB}}   & \textbf{EarlyRobust}               &$91.8$ & $\mathbf{\underline{63.8}}$   \\
 & \quad \textbf{w/o} Adv   & $\mathbf{\underline{92.0}}$  & $6.4$   \\ 
 & \quad \textbf{w/o} Regularizer                   & $87.5$     & $31.2$  \\
  & \quad  {Random Tickets}                   & $91.1$     & $6.2$   \\ \hline
 
\multirow{4}{*}{\textbf{SST-2}} & \textbf{EarlyRobust}              & $\mathbf{\underline{92.8}}$     & $\mathbf{\underline{28.8}}$   \\ 
& \quad \textbf{w/o} Adv                      &  $92.2$    & $13.9$   \\ 
& \quad \textbf{w/o} Regularizer                  & $91.3$     & $8.5$   \\ 
  & \quad  {Random Tickets}                   & $90.9$     & $9.1$   \\ \hline

\multirow{4}{*}{\textbf{AG NEWS}}   & \textbf{EarlyRobust}               & 94.4     & $\mathbf{\underline{41.0}}$   \\ 
& \quad \textbf{w/o} Adv & $\mathbf{\underline{94.6}}$ & $31.0$   \\ 
& \quad \textbf{w/o} Regularizer   & $92.5$  & $31.2$ \\
  & \quad  {Random Tickets}                  & $94.0$     & $19.4$   \\ 
  \hline\hline

\end{tabular}
\caption{Ablation study. \textbf{Aua$\%$} is obtained after using TextFooler attack. Results of randomly pruned tickets are the average of 5 trials with different random seeds. 
Both Adversarial loss objective and sparsity-inducing regularizer play a fundamental role for high-quality tickets. Random tickets suffer a large degradation in robustness, indicating that robust tickets are non-trivial. 
}
\label{tab:ablation}
\end{table}
\subsection{Different Early-stopping Thresholds}
\begin{figure}[t]
    \centering
    \includegraphics[width=0.7\linewidth]{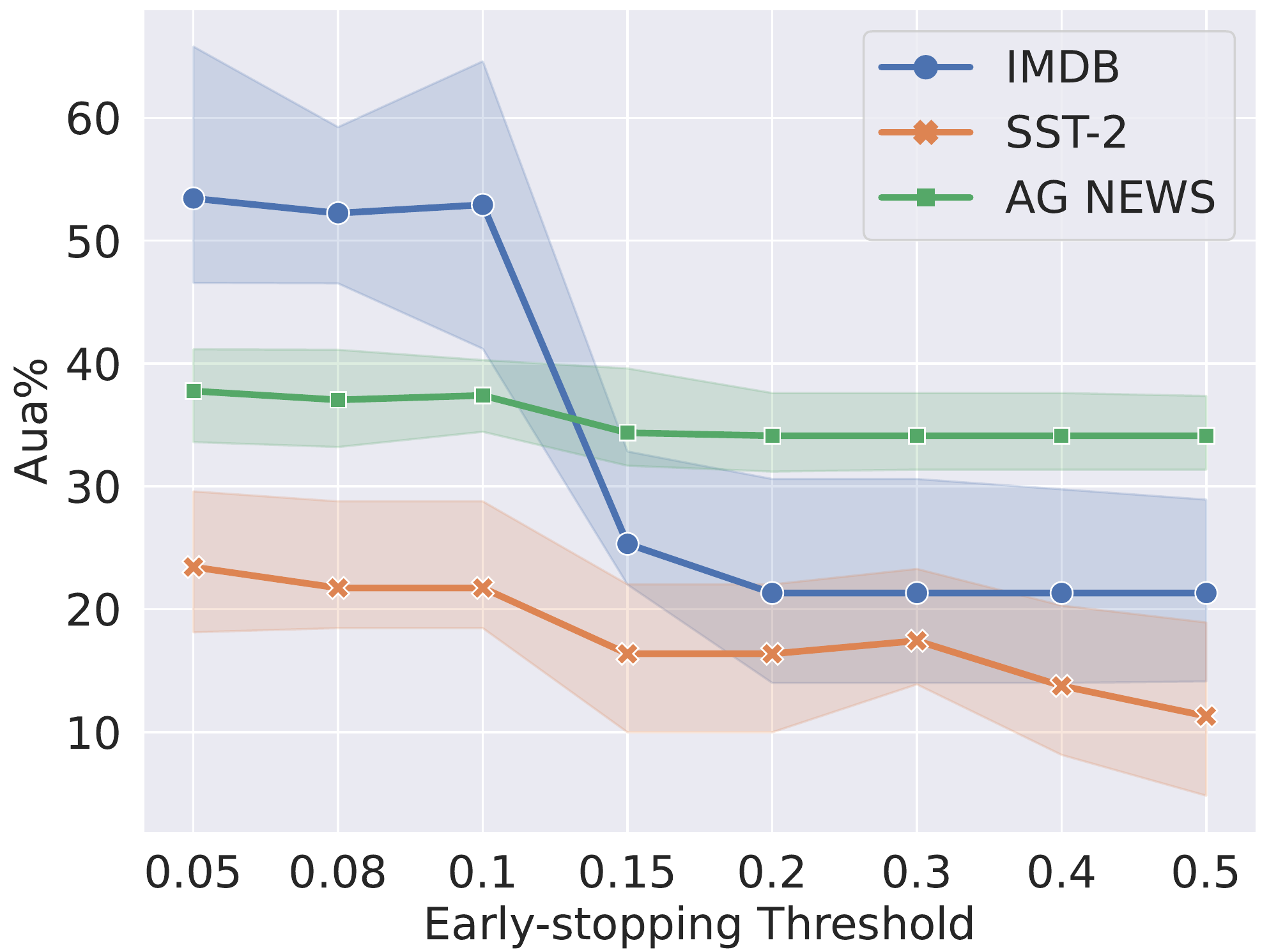}
	\caption{Influence of different early-stopping thresholds (i.e., $\gamma$) on model robustness. \textbf{Aua$\%$} is obtained after using TextFooler attack and the results are obtained from 5 trials of different random seeds. Model robustness converges when $\gamma$ decreases to around $0.1$.}
	\label{fig:Different epsilon}
\end{figure}
Here we study how different early-stopping thresholds affect the model's robustness. Results in Figure \ref{fig:Different epsilon} show that the model's robustness converges when $\gamma$ is reduced to around $0.1$. Considering training efficiency, we think $0.1$ is a reasonable threshold.

\subsection{Importance of EarlyRobust Tickets Initialization and Structure} \label{Sec:The Importance of EarlyRobust Tickets Initialization and Structure}
\begin{table}[t]
\renewcommand\arraystretch{1.15}
\setlength\tabcolsep{5pt}
\centering
\small
\begin{tabular}{l|lcc}
\hline
\hline
\textbf{Dataset}   &\textbf{Method} & \textbf{Clean$\%$} & \textbf{Aua$\%$} \\ \hline
 \multirow{4}{*}{\textbf{IMDB}}   & \textbf{EarlyRobust}               &$\mathbf{\underline{91.8}}$ & $\mathbf{\underline{63.8}}$   \\
 & \quad \textbf{w/o} Initialization   & $82.4$  & $0.0$   \\ 
 & \quad \textbf{w/o} Structure                   & $89.2$     & $4.0$   \\ \hline
\multirow{4}{*}{\textbf{SST-2}} & \textbf{EarlyRobust}              & $\mathbf{\underline{92.8}}$     & $\mathbf{\underline{28.8}}$   \\ 
& \quad \textbf{w/o} Initialization                      &  $81.5$    & $0.8$   \\ 
& \quad \textbf{w/o} Structure                  & $88.4$     & $20.0$   \\ \hline
\multirow{4}{*}{\textbf{AG NEWS}}   & \textbf{EarlyRobust}               & $\mathbf{\underline{94.4}}$     & $\mathbf{\underline{41.0}}$   \\ 
& \quad \textbf{w/o} Initialization & $92.2$ & $0.6$   \\ 
& \quad \textbf{w/o} Structure   & $\mathbf{\underline{94.4}}$  & $36.6$ \\ 
\hline
\hline
\end{tabular}
\caption{Importance of EarlyRobust ticket initialization and structure. \textbf{Aua$\%$} is obtained under TextFooler attack. 
The pre-trained initialization and the structure are both indispensable for EarlyRobust tickets.
The pre-trained initialization seems more important than the structure. 
}
\label{tab:the importance of init and struct}
\end{table}
According to LTH, the winning tickets can not be trained effectively without original initialization and the ticket structure also plays a role \cite{DBLP:conf/iclr/FrankleC19}. 
Therefore, we follow a similar way in \cite{zheng-etal-2022-robust} to study the importance of the initialization and the structure of EarlyRobust tickets. 
Specifically, we re-initialize the weights of EarlyRobust tickets to exclude the effect of the initialization.
We employ the full BERT and re-initialize the weights which are not contained in the robust tickets to exclude the effect of the structure while preserving the effect of the initialization.
Table \ref{tab:the importance of init and struct} demonstrates that the pre-trained initialization and the structure are both fundamental for model performance. 
Another observation is that without the initialization, robust tickets suffer a greater performance drop than without the structure.

\subsection{Impact of Sparsity}
\begin{figure}[t]
    \small
	\centering
	\subfigure[Clean\% on FC Sparsity]{
        \begin{minipage}[t]{0.48\linewidth}
        \centering
        \includegraphics[width=0.9\linewidth]{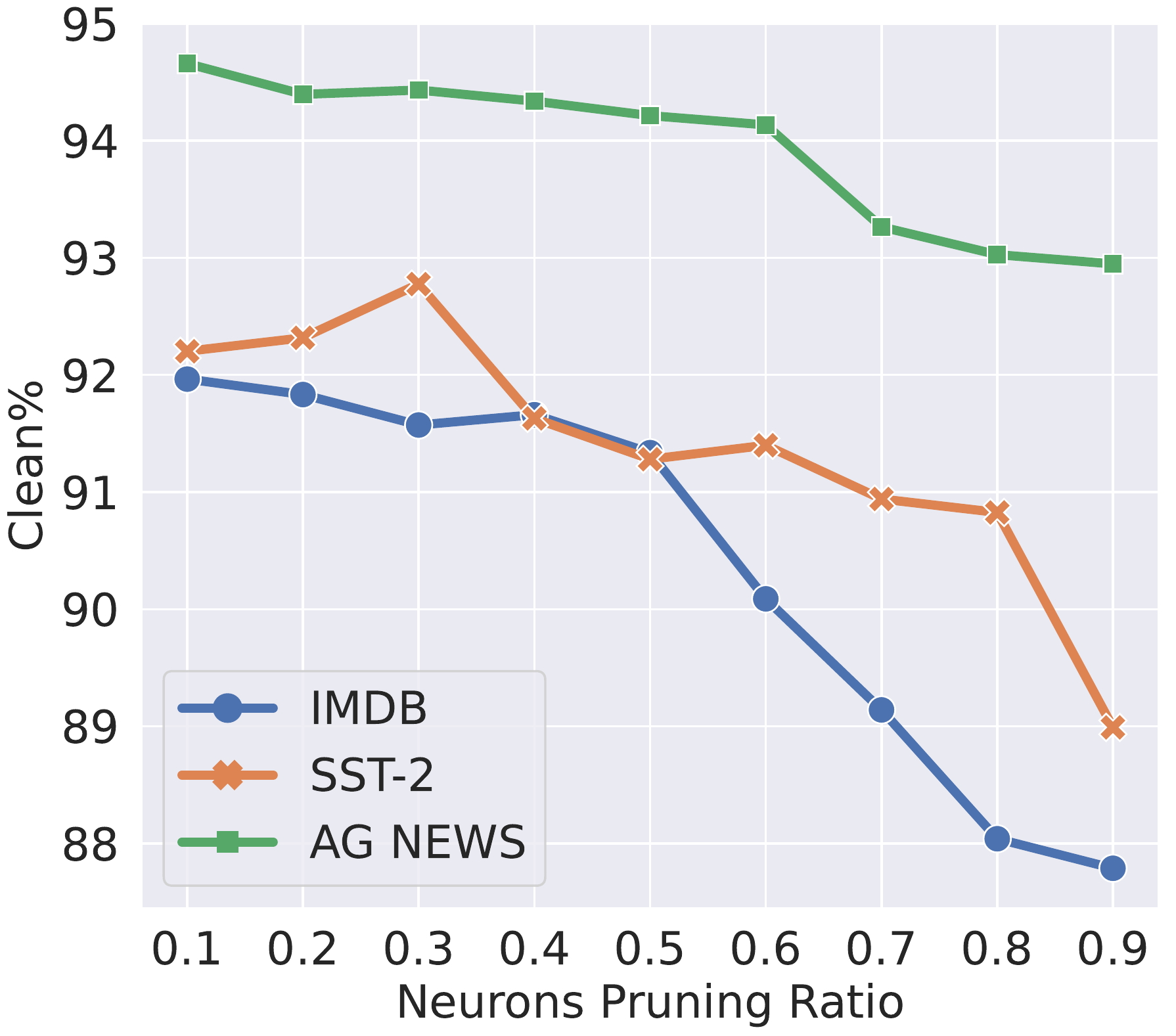}
        \label{Impact of FC Pruning Ratio on Clean}
        \end{minipage}%
    }%
	\centering
	\subfigure[Clean\% on Heads Sparsity]{
        \begin{minipage}[t]{0.48\linewidth}
        \centering
        \includegraphics[width=0.9\linewidth]{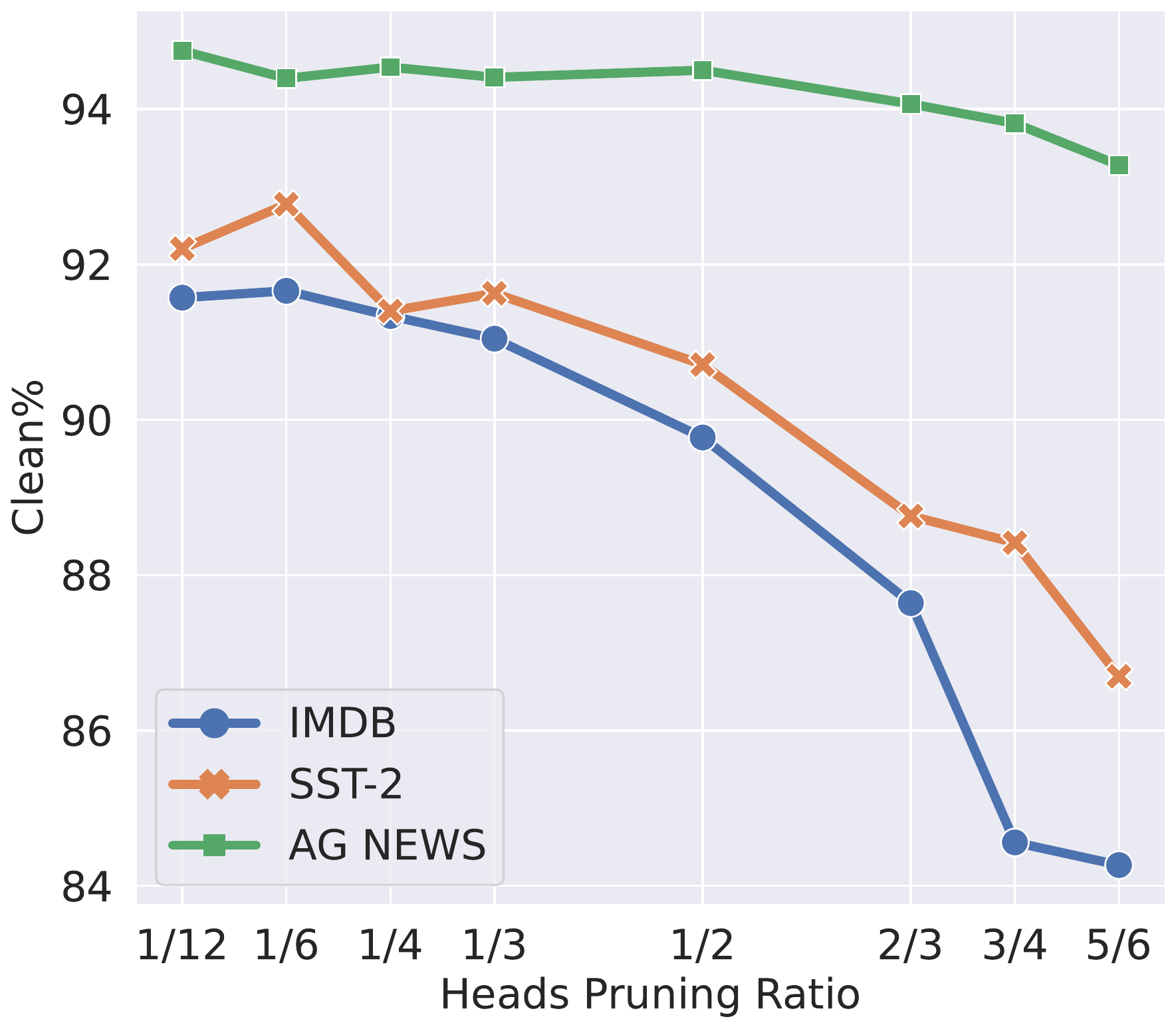}
        \label{Impact of Heads Pruning Ratio on Clean}
        \end{minipage}%
    }%
	
	\centering
	\subfigure[Aua\% on FC Sparsity]{
        \begin{minipage}[t]{0.48\linewidth}
        \centering
        \includegraphics[width=0.9\linewidth]{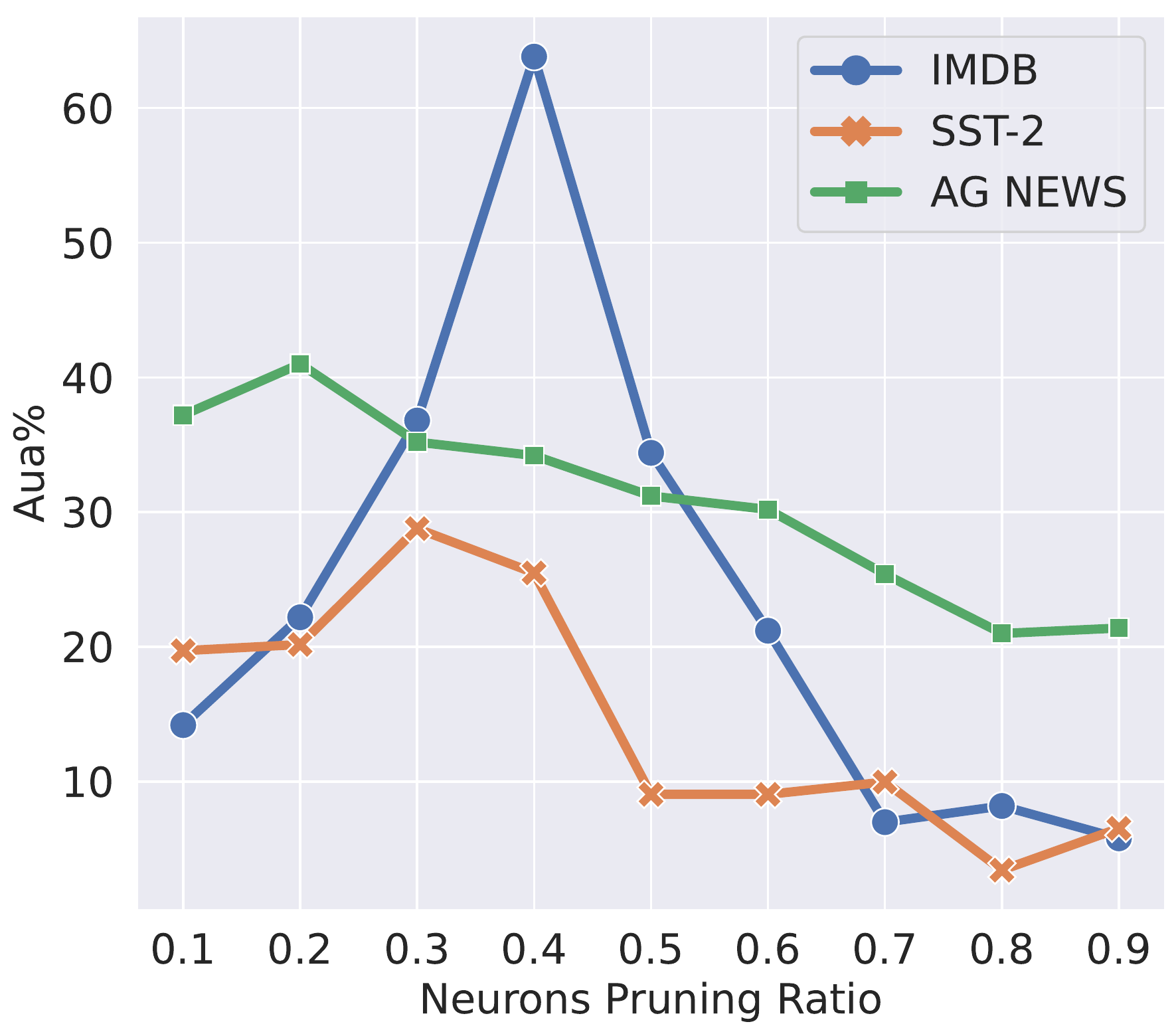}
        \label{Impact of FC Pruning Ratio on Aua}
        \end{minipage}%
    }%
	\centering
	\subfigure[Aua\% on Heads Sparsity]{
        \begin{minipage}[t]{0.48\linewidth}
        \centering
        \includegraphics[width=0.9\linewidth]{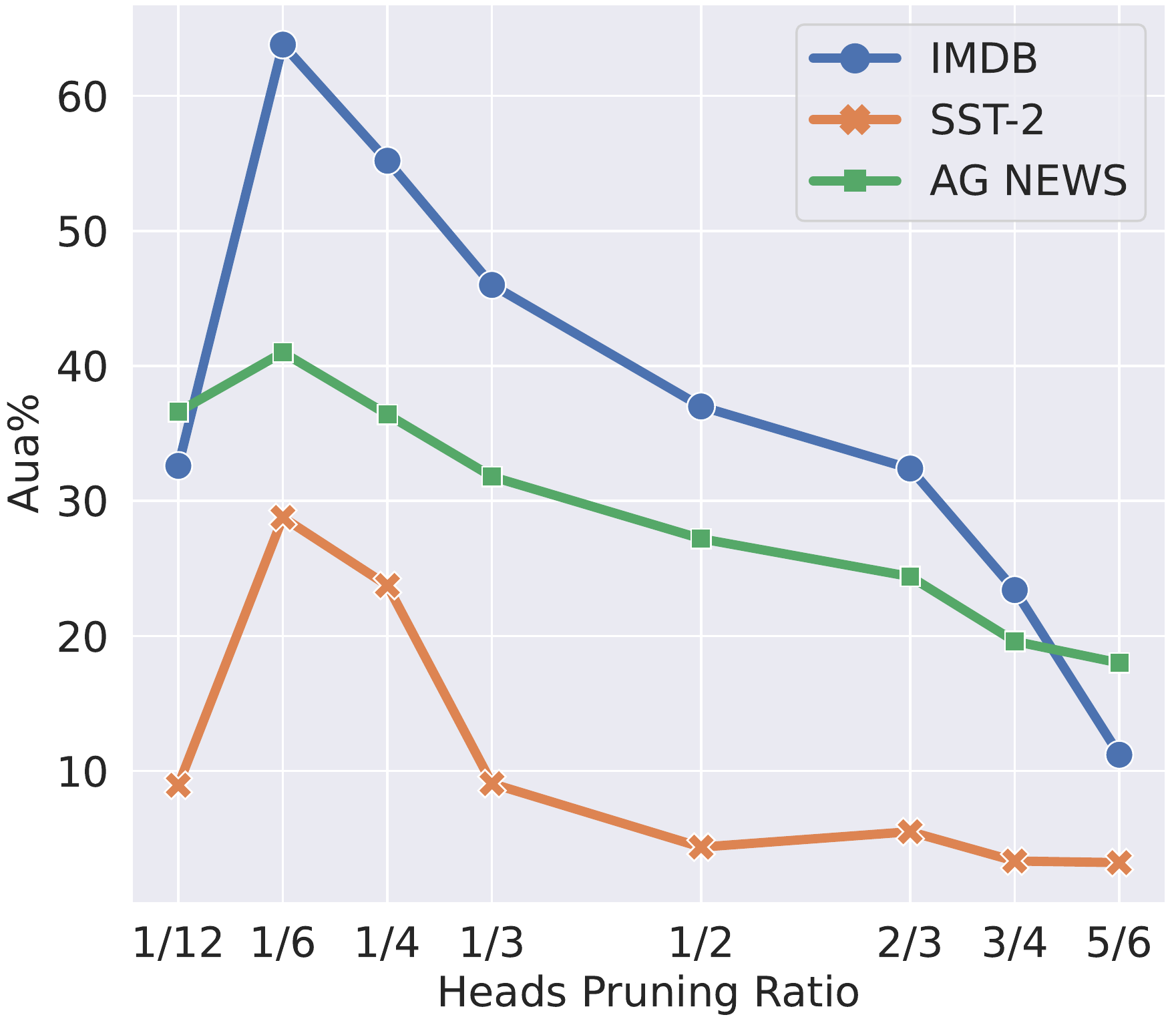}
        \label{Impact of Heads Pruning Ratio on Aua}
        \end{minipage}%
    }%

	\caption{Impact of Sparsity on \textbf{Clean$\%$} and \textbf{Aua$\%$}. \textbf{Aua$\%$} is obtained under TextFooler attack.}
	\label{fig:Impact of Sparsity on EarlyRobust Tickets}
\end{figure}

In our method, structured sparsity can not only bring speedups but also save the storage. 
Figure \ref{fig:Impact of Sparsity on EarlyRobust Tickets} illustrates that as the sparsity (both for heads and neurons) increases, the accuracy shows a gradually decreasing trend, but the robustness is different. The robustness first increases and then degrades sharply as sparsity grows, which is similar to the observations in \cite{DBLP:conf/nips/FuYZWOCL21}. 
To extract high-quality tickets, $1/6 \sim 1/4$ and $20\% \sim 40\%$ are suitable pruning ratio ranges for heads and neurons, respectively.
We also visualize the sparsity patterns in Appendix \ref{Appendix:Sparsity Pattern}.

\subsection{Global Pruning vs. Layer-wise Pruning} \label{sec:Global Pruning vs. Layer-wise Pruning}

\begin{table}[t]
\renewcommand\arraystretch{1.2}
\setlength\tabcolsep{4pt}
\centering
\small

\begin{tabular}{ccc}
	\hline
	\hline
	\textbf{Clean\%} & \multirow{2}{*}{\textbf{Heads-L}} & \multirow{2}{*}{\textbf{Heads-G}}  \\ \cline{1-1} 
	\textbf{Aua\%}
	\\ 
	
	\hline
	
	\multirow{2}{*}{\textbf{Neurons-L}}   & $91.5$               &$91.6$   \\ \cline{2-3}
	& $1.2$               &$35.4$   \\
	
	\hline
	
	\multirow{2}{*}{\textbf{Neurons-G}}   & $91.4$              & $\mathbf{\underline{91.7}}$       \\ \cline{2-3}
	& $56.0$               &$\mathbf{\underline{63.8}}$   \\
	
	\hline\hline
	
\end{tabular}

\caption{Influence of global and layer-wise pruning on model performance on IMDB. 
\textbf{-L} means layer-wise while \textbf{-G} means global. 
\textbf{Aua$\%$} is obtained under TextFooler attack.
Global pruning outperforms layer-wise pruning in terms of robustness.
}
\label{tab:Influence of global and layer-wise pruning}
\end{table}

As mentioned in Sec.\ref{sec:Drawing and Fine-tuning Structured Robust tickets}, we prune self-attention heads and intermediate neurons separately with global approach.
Here we study how global pruning and layer-wise pruning influence model performance. Table \ref{tab:Influence of global and layer-wise pruning} demonstrates the results on IMDB.\footnote{See Appendix \ref{appendix:More Results of Global Pruning vs. Layer-wise Pruning} for more results.} We can observe that: 1) Layer-wise pruning performs as well as global pruning in accuracy, but it induces a degradation in robustness, revealing that different layers have different degrees of contribution to robustness. So different layers should not be pruned with the same ratio. 
2) The model's robustness drops more dramatically when we use layer-wise pruning for neurons than for heads,
suggesting that intermediate neurons are more sensitive to pruning approaches.

\section{Conclusion}
In this paper, we delve into the optimization process of adversarial training on BERT, and observe the early convergence of robust connectivity patterns. Based on this, we propose EarlyRobust, a method that shapes structured robust early-bird tickets under the guidance of the adversarial loss objective and then utilizes efficient fine-tuning to train them. 
Experiments on various datasets demonstrate that our method achieves competitive or even better robustness compared to other strong baselines while maintaining a very low computational and storage consumption.

\section*{Limitation}
In this work, we propose a method to draw structured robust early-bird tickets, which can be used as an efficient alternative to adversarial training. However, there are still limitations to be explored in future work: 
1) Model robustness drops sharply when the pruning ratio is too large, so our robust tickets do not have high sparsity. We expect to compress the robust early-bird tickets to a smaller size while maintaining a high quality in the future. 
2) We draw the robust early-bird tickets from pre-trained models, and future work includes searching for robust early-bird tickets from randomly initialized networks.
\section*{Ethics Statement}
Our method is proposed for efficient adversarial training, so it is environmental-friendly. It achieves up to $7\times \sim 13\times$ speedups compared to standard adversarial training, saving computational resources by a significant margin. Moreover, the proposed robust early-bird tickets with structured sparsity bring savings in storage, so it is easy to deploy them on mobile devices and edge devices.
\section*{Acknowledgements}
The authors wish to thank the anonymous reviewers for their helpful comments. This work was partially funded by National Natural Science Foundation of China (No. 62206057, 61906176, 62076069, 61976056), Program of Shanghai Academic Research Leader (No. 22XD1401100), Beijing Academy of Artificial Intelligence (BAAI) and CCF-Tencent Open Fund.
\bibliography{anthology,custom,main}
\bibliographystyle{acl_natbib}

\appendix

\clearpage
\section{Appendix}
\subsection{Solving the Inner Max Problem of Adversarial Loss Objective} \label{appendix:Solving the Inner Max Problem of Adversarial Loss Objective}
We solve the inner max problem(generate the worst-case perturbation, i.e., $\mathcal{L}_{adv}(\theta,c)$) with K-step projected gradient descent strategy \cite{DBLP:conf/iclr/MadryMSTV18}, and the ($k+1$)-th perturbation is defined as :
\begin{equation}
\begin{aligned}
\delta_{k+1} = \prod_{\lVert \delta \rVert \leq \epsilon}\left (\delta_k + \eta\frac {g(\delta_k)}{\lVert g(\delta_k) \rVert}  \right ), \label{inner_max}
\end{aligned}
\end{equation}
where $g(\delta_k) = \nabla_{\delta_k} \mathcal{L}(f(x+\delta_k;\theta,c),y)$ is the gradient of the loss with respect to $\delta$, and $\prod_{\lVert \delta \rVert \leq\epsilon}(\cdot)$ performs a projection onto the $\epsilon$-ball, which is a Frobenius normalization ball.

\subsection{Implementation Details} \label{Appendix:Implementation Details}
\subsubsection{Details for Adversarial Attack}
We adopt textattack \cite{morris2020textattack} to implement adversarial attack methods. \textbf{Aua\%} and \textbf{\#Query} are evaluated on all 872 test examples for SST-2, 500 randomly selected test samples for IMDB, AG NEWS, QNLI and QQP. For BERT-Attack, we set the neighbor vocabulary size to $50$, and set the sentence similarity to $0.2$; for the other two attack methods, we use the default parameters of third-party libraries.
\subsubsection{Hardware Details}
Experiments on SST-2, QNLI and QQP are run on seven 1080Ti GPUs and the sequence length is set to 128, while experiments on AG NEWS and IMDB are run on eight 2080Ti GPUs and the sequence length is set to 256.
\subsubsection{Hyperparameters}
There are four hyperparameters introduced by the adversarial loss objective: the initial magnitude of perturbations $\epsilon_0$, the number of gradient descent steps for adversary $s$, the perturbation step size $\tau$, and we do not constrain the bound of perturbations. We report the learning rate $\eta$ and the learning rate is the same at both the searching and fine-tuning stages. Moreover, we report the regularization penalties $\lambda_{\rm  \scriptscriptstyle H}$ and $\lambda_{\rm \scriptscriptstyle F}$. We list the hyperparameters used for each task in Table \ref{tab:hyperparameters}.

\begin{table}[H]
\renewcommand\arraystretch{1.1}
\setlength\tabcolsep{5pt}
\centering
\small

\begin{tabular}{c|cccccc}
\hline
\hline
\multicolumn{1}{c|}{\textbf{Datasets}} &
\multicolumn{1}{c}{\textbf{$\epsilon_0$}} &
\multicolumn{1}{c}{\textbf{$s$}} &
\multicolumn{1}{c}{\textbf{$\tau$}} &
\multicolumn{1}{c}{\textbf{$\eta$}} &
\multicolumn{1}{c}{\textbf{$\lambda_{\rm  \scriptscriptstyle H}$}} &
\multicolumn{1}{c}{\textbf{$\lambda_{\rm  \scriptscriptstyle F}$}}\\
\hline
\textbf{IMDB} & $0.05$ &$5$& $0.01$& $2e-5$& $1e-5$ & $2e-4$\\
\textbf{SST2} & $0.05$ &$5$& $0.02$& $2e-5$& $1e-4$ & $3e-4$\\
\textbf{AGNEWS} & $0.05$ &$5$& $0.01$& $2e-5$& $3e-4$ & $5e-5$\\
\textbf{QNLI} & $0.05$ &$5$& $0.01$& $2e-5$& $2e-4$ & $3e-4$\\
\textbf{QQP} & $0.05$ &$5$& $0.01$& $2e-5$& $1e-5$ & $1e-5$\\

\hline
\hline
\end{tabular}
\caption{Hyperparameters in the proposed method.}
\label{tab:hyperparameters}
\end{table}

\subsection{Sparsity Pattern}\label{Appendix:Sparsity Pattern}
\begin{figure*}[t]
	\centering
	\subfigure[IMDB]{
        \begin{minipage}[t]{0.32\linewidth}
        \centering
        \includegraphics[width=1\linewidth]{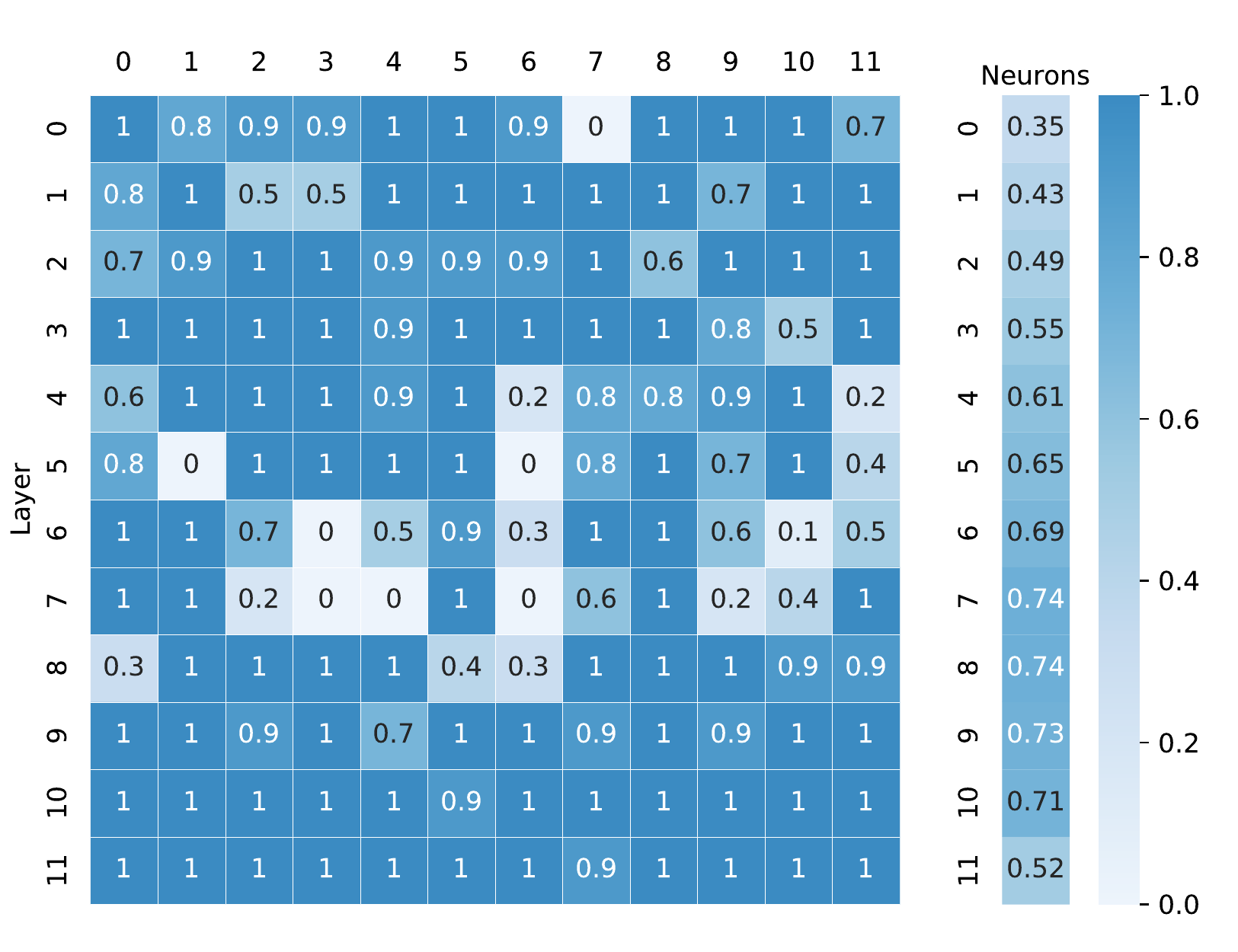} 
        \label{IMDB}
        \end{minipage}%
    }%
	\centering
	\subfigure[SST2]{
        \begin{minipage}[t]{0.32\linewidth}
        \centering
        \includegraphics[width=1\linewidth]{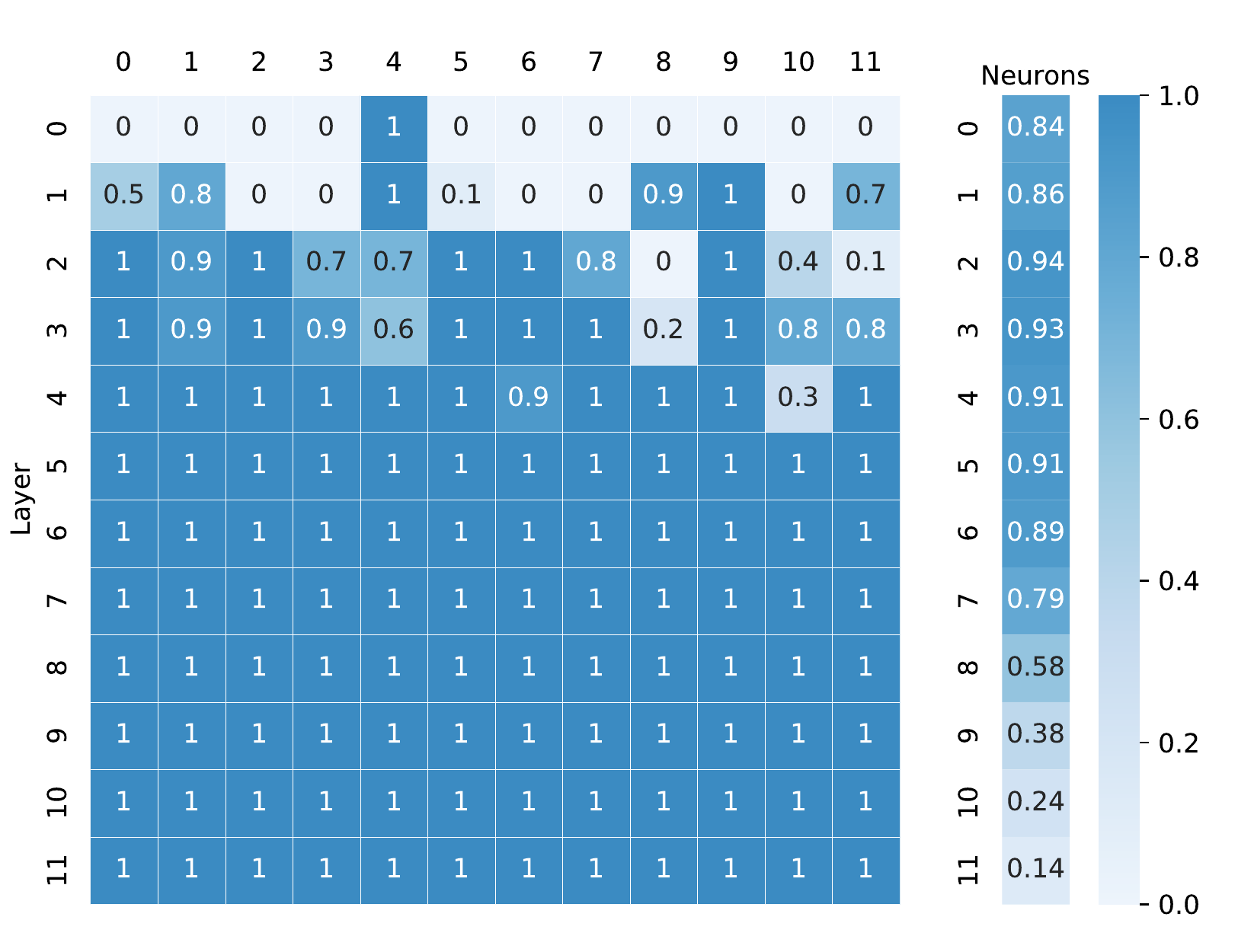}
        \label{IMDB}
        \end{minipage}%
    }%
	\centering
	\subfigure[AG NEWS]{
        \begin{minipage}[t]{0.32\linewidth}
        \centering
        \includegraphics[width=1\linewidth]{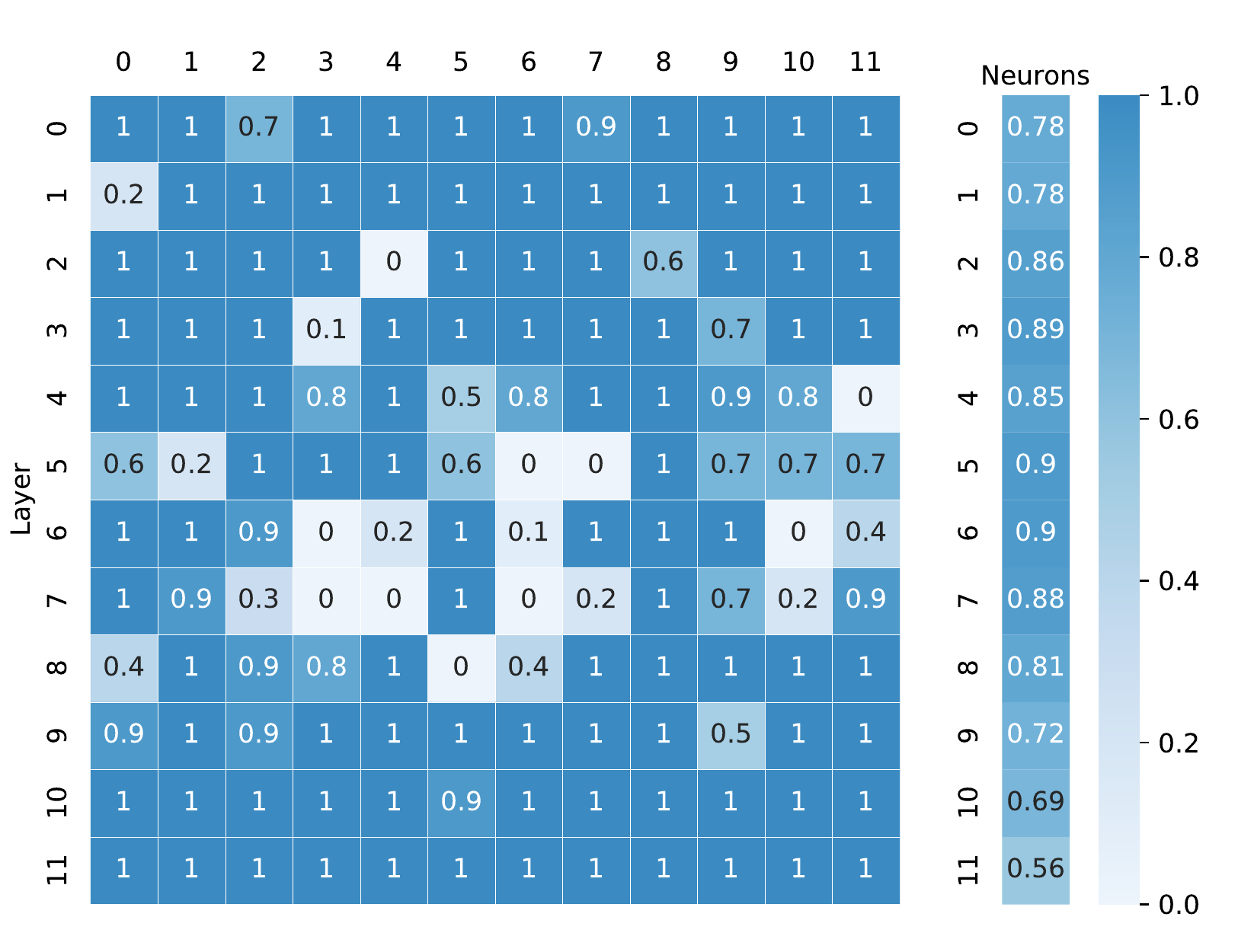}
        \label{IMDB}
        \end{minipage}%
    }%

	\caption{Sparsity pattern of EarlyRobust tickets. For each figure, cells in the left give the average number of random seeds where a given head survives; cells in the right give the average percentage of surviving weights across 10 random seeds. We can observe that the sparsity pattern is dataset-dependent and interestingly, on a given dataset, the pruning for heads and intermediate neurons is concentrated in different layers.}
	\label{fig:Sparsity pattern}
\end{figure*}
Figure \ref{fig:Sparsity pattern} illustrates the sparsity pattern of robust tickets on three datasets. We can observe that for different datasets, the sparsity pattern varies greatly. So we speculate that it is dataset-dependent. Another interesting finding is that on a given dataset, the pruning for attention heads and intermediate neurons is concentrated in different layers, i.e., if attention heads are pruned centrally in some layers, the pruning on intermediate neurons is concentrated in other layers. For example, on SST-2, our method prunes heads mainly in bottom layers, while pruning intermediate neurons mainly in top layers.

\subsection{More Results of Global Pruning vs. Layer-wise Pruning} \label{appendix:More Results of Global Pruning vs. Layer-wise Pruning}

Table \ref{tab:Global Pruning vs. Layer-wise Pruning on SST-2.} and Table \ref{tab:Global Pruning vs. Layer-wise Pruning on AG NEWS.} show the results on SST-2 and AG NEWS, respectively. We can consistently find that \emph{global} pruning performs better than \emph{layer-wise} pruning in terms of robustness.

\begin{table}[H]
\renewcommand\arraystretch{1.4}
\setlength\tabcolsep{4pt}
\centering
\small

\begin{tabular}{ccc}
	\hline
	\hline
	\textbf{Clean\%} & \multirow{2}{*}{\textbf{Heads-L}} & \multirow{2}{*}{\textbf{Heads-G}}  \\ \cline{1-1} 
	\textbf{Aua\%}
	\\ 
	
	\hline
	
	\multirow{2}{*}{\textbf{Neurons-L}}   & $90.1$               &$91.5$   \\ \cline{2-3}
	& $11.7$               &$12.0$   \\
	
	\hline
	
	\multirow{2}{*}{\textbf{Neurons-G}}   & $91.6$              & $\mathbf{\underline{92.8}}$       \\ \cline{2-3}
	& $24.3$               &$\mathbf{\underline{28.8}}$   \\
	
	\hline\hline
	
\end{tabular}

\caption{Global Pruning vs. Layer-wise Pruning on SST-2.}
\label{tab:Global Pruning vs. Layer-wise Pruning on SST-2.}
\end{table}

\begin{table}[H]
\renewcommand\arraystretch{1.4}
\setlength\tabcolsep{4pt}
\centering
\small
\begin{tabular}{ccc}
	\hline
	\hline
	\textbf{Clean\%} & \multirow{2}{*}{\textbf{Heads-L}} & \multirow{2}{*}{\textbf{Heads-G}}  \\ \cline{1-1} 
	\textbf{Aua\%}
	\\ 
	
	\hline
	
	\multirow{2}{*}{\textbf{Neurons-L}}   & $94.5$               &$\mathbf{\underline{94.6}}$   \\ \cline{2-3}
	& $36.8$               &$38.4$   \\
	
	\hline
	
	\multirow{2}{*}{\textbf{Neurons-G}}   & $94.4$              & $94.5$       \\ \cline{2-3}
	& $39.6$               &$\mathbf{\underline{41.0}}$   \\
	
	\hline\hline
	
\end{tabular}

\caption{Global Pruning vs. Layer-wise Pruning on AG NEWS.}
\label{tab:Global Pruning vs. Layer-wise Pruning on AG NEWS.}
\end{table}
\label{sec:appendix}

\end{document}